\documentclass[conference]{IEEEtran}
\IEEEoverridecommandlockouts
\usepackage{multirow}
\usepackage{cite}
\usepackage{amsmath,amssymb,amsfonts}
\usepackage{algorithmic}
\usepackage{graphicx}
\usepackage{textcomp}
\usepackage{xcolor}
\usepackage{balance}
\usepackage{url}
\usepackage{booktabs}
\usepackage{multicol}
\usepackage{subcaption}
\graphicspath{{figures/}}

\def\BibTeX{{\rm B\kern-.05em{\sc i\kern-.025em b}\kern-.08em
    T\kern-.1667em\lower.7ex\hbox{E}\kern-.125emX}}

\begin{document}

\title{An Evaluation of Low Overhead Time Series Preprocessing Techniques for Downstream Machine Learning
    \thanks{
    Research was sponsored by the United States Air Force Research Laboratory and the United States Air Force Artificial Intelligence Accelerator and was accomplished under Cooperative Agreement Number FA8750-19-2-1000. The views and conclusions contained in this document are those of the authors and should not be interpreted as representing the official policies, either expressed or implied, of the United States Air Force or the U.S. Government. The U.S. Government is authorized to reproduce and distribute reprints for Government purposes notwithstanding any copyright notation herein.
    }
}

\author{
        Matthew L. Weiss\IEEEauthorrefmark{1},
        Joseph McDonald\IEEEauthorrefmark{1},
        David Bestor\IEEEauthorrefmark{1},
        Charles Yee\IEEEauthorrefmark{1},
        Daniel Edelman\IEEEauthorrefmark{2}, \\
        Michael Jones\IEEEauthorrefmark{1},
        Andrew Prout\IEEEauthorrefmark{1},
        Andrew Bowne\IEEEauthorrefmark{4},
        Lindsey McEvoy\IEEEauthorrefmark{4}, 
        Vijay Gadepally\IEEEauthorrefmark{1},
        Siddharth Samsi\IEEEauthorrefmark{1} \\
    \IEEEauthorrefmark{1} MIT Lincoln Laboratory,
    \IEEEauthorrefmark{2} MIT,  
    \IEEEauthorrefmark{4} US Air Force}

\maketitle

\noindent \copyright{ 2022 IEEE.} Personal use of this material is permitted. Permission from IEEE must be obtained for all other uses, in any current or future media, including reprinting/republishing this material for advertising or promotional purposes, creating new collective works, for resale or redistribution to servers or lists, or reuse of any copyrighted component of this work in other works.

\begingroup\renewcommand\thefootnote{\textsection}
\footnotetext{Corresponding author. Email : \url{mlweiss@ll.mit.edu}}
\endgroup

\begin{abstract}
In this paper we address the application of preprocessing techniques to multi-channel time series data with varying lengths, which we refer to as the alignment problem, for downstream machine learning.  The misalignment of multi-channel time series data may occur for a variety of reasons, such as missing data, varying sampling rates, or inconsistent collection times.  We consider multi-channel time series data collected from the MIT SuperCloud High Performance Computing (HPC) center, where different job start times and varying run times of HPC jobs result in misaligned data. This misalignment makes it challenging to build AI/ML approaches for tasks such as compute workload classification. Building on previous supervised classification work with the MIT SuperCloud Dataset, we address the alignment problem via three broad, low overhead approaches: sampling a fixed subset from a full time series, performing summary statistics on a full time series, and sampling a subset of coefficients from time series mapped to the frequency domain.  Our best performing models achieve a classification accuracy greater than 95\%, outperforming previous approaches to multi-channel time series classification with the MIT SuperCloud Dataset by 5\%.  These results indicate our low overhead approaches to solving the alignment problem, in conjunction with standard machine learning techniques, are able to achieve high levels of classification accuracy, and serve as a baseline for future approaches to addressing the alignment problem, such as kernel methods.
\end{abstract}
\section{Introduction}\label{sec:introduction}

Time series are ubiquitous in many domains such as medicine, speech, finance, control systems, and computing to name a few. A relatively new area of multi-modal time-series is the modern cloud or High Performance Computing (HPC) system, where time series data can arise from a large variety of sources. These sources include highly granular data collected from compute infrastructure such as the CPUs/GPUs, networking, and file system utilization to higher level time series data such as overall cluster utilization as a function of time. These data are an important source of monitoring datacenter operations and can provide actionable insights into overall system health and operating efficiency. Additionally, environmental data from the datacenter such as temperature, humidity, and power consumption are critical for monitoring and operating datacenters from the perspective of energy efficiency and physical factors that can drastically affect system operation. This area of time series analysis is a ripe target for the application of AI/ML approaches for optimization. For example, Google researchers applied machine learning to optimize datacenter cooling~\cite{cooling2018,deepmind}. However, neither the data nor models are publicly available. 

To enable the development of AI/ML models in this domain, we have released the MIT SuperCloud Dataset~\cite{dcc}, which consists of a large collection of time series data collected from an operational supercomputing center. This dataset consists of time series of hardware utilization which includes the CPU and GPU utilization over time, memory usage, file I/O, and other signals monitored on the system.  The MIT SuperCloud Dataset is freely available for download via AWS Open Data Registry.  Detailed instructions for downloading the data are available at \url{https://dcc.mit.edu/data}.

The goal of the MIT SuperCloud Dataset is to foster the development of AI/ML approaches for improving cluster operations. One important application includes the MIT SuperCloud Workload Classification Challenge~\cite{wcc}, which aims to use machine learning to identify compute workloads on the system using time series data collected from all jobs running on the system.  This paper is a further contribution to the MIT SuperCloud Workload Classification Challenge, extending the baseline implementations and results presented in~\cite{wcc}.

To effectively perform machine learning tasks, such as classification or regression, on time series data certain requirements are necessary of the data.  One such requirement is that the feature space of the time series align or are of the same dimension.  This is a general principle of machine learning algorithms which require an input matrix where the rows represent the trials or samples and the columns correspond to the feature space.  However, in the case of real-world time series, ensuring that data occupy the same dimension or feature space, which we refer to as \textit{alignment} herein, is often non-trivial.  Potential causes of misalignment may include missing data, time series which are collected over different time intervals, or data collected at different sampling rates. In this work, which is part of the larger MIT SuperCloud Datacenter Challenge~\cite{dcc}, we build on the results in~\cite{wcc} where we address the problem of classifying non-aligned time series data collected from the MIT SuperCloud High Performance Computing Center~\cite{wcc}.

\subsection{Prior Work}\label{ssec:prior_work}
A variety of approaches to the time series alignment problem exist, ranging from the padding or interpolating of missing values to the application of state-of-the-art neural networks.  While padding and interpolation come with a low computational overhead, they result in all time series having the length of the largest time series.  In order not to significantly grow the size of the time series data, other methods based on dynamic time warping \cite{cohen2021aligning,cuturi2017soft,sakoe1978dynamic,vayer2020time} and Support Vector Machines/Kernel methods \cite{bagheri2016support,badiane2018kernel,shimodaira2001dynamic} have also been suggested.  Recent attempts also exist that leverage neural networks \cite{horak2019comparison,iwana2020dtw,kurumatani2018time,zhang2021graph}, which are able to learn complex patterns from data but at the cost of the high computational complexity required by neural networks. 

\subsection{Our Contribution}\label{ssec:out_contribution}
In this work we avoid both dataset augmentation through padding/interpolation and the higher overhead computations discussed above, while achieving high classification accuracy, by employing low computational overhead approaches to the time series alignment problem.  Specifically,
\begin{itemize}
\item Establish a baseline for improved classification accuracy by sampling N points from each time series.
\item Leverage low overhead preprocessing techniques, based upon summary statistics and the Fourier Transform, to generate N sample time series which consider the full time series.
\item Demonstrate these low overhead time series preprocessing techniques achieve upwards of 95\% accuracy using well-known machine learning algorithms, an improvement of 5\% over previous work based upon sampling N contiguous point in a time series.
\end{itemize}
\begin{figure*}[h]
    \centering
    \begin{subfigure}[t]{\textwidth}
        \centering 
        \includegraphics[scale=0.4]{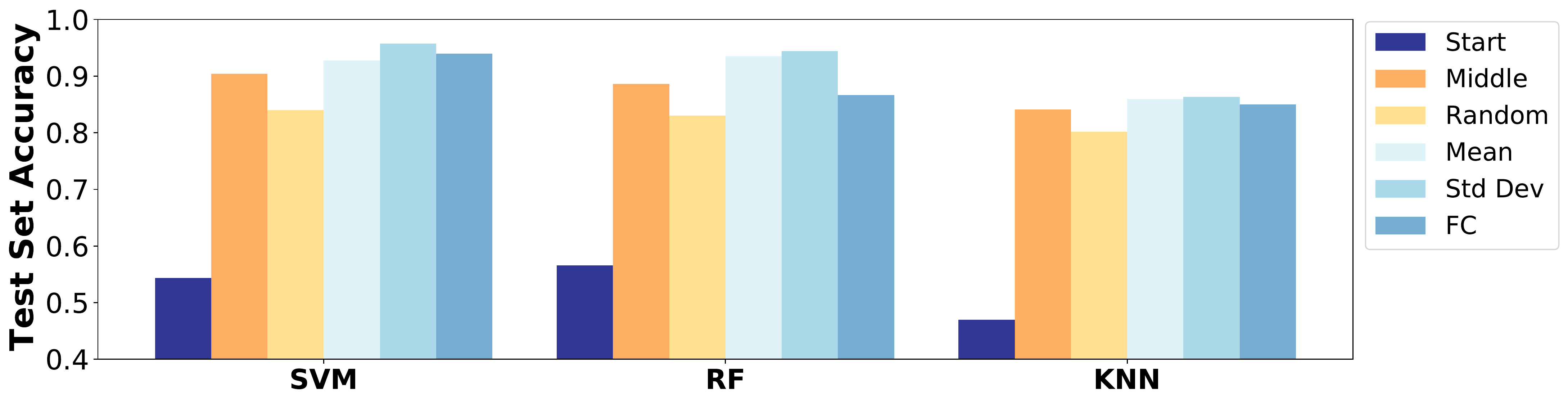} 
        \caption{N=100}
        \label{fig:bar_100}
    \end{subfigure}
    \\
    \begin{subfigure}[t]{\textwidth}
        \centering    
        \includegraphics[scale=0.4]{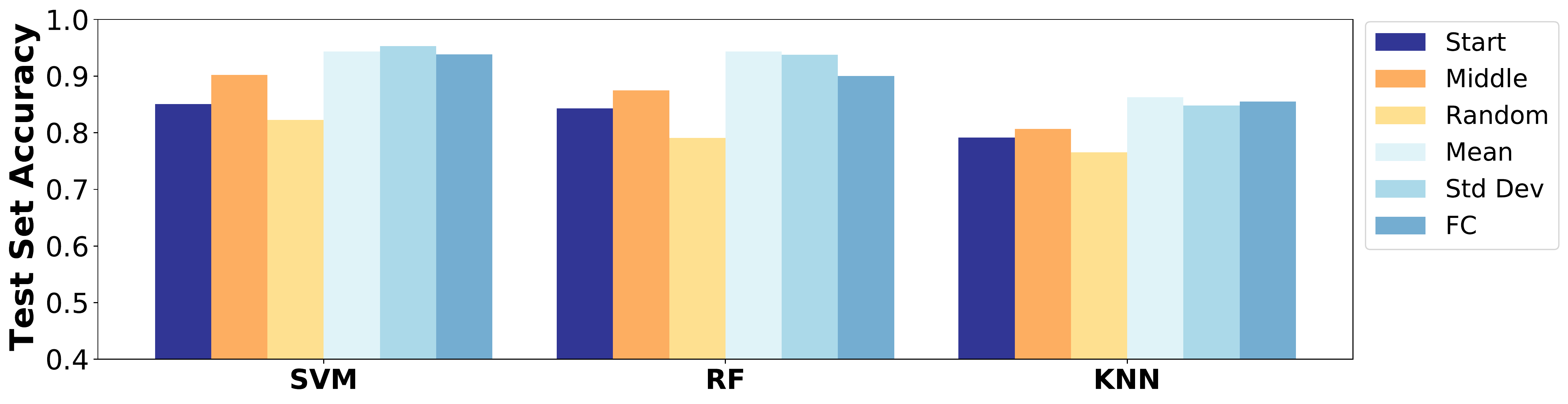} 
        \caption{N=1000}
        \label{fig:1000_std}
    \end{subfigure}
    \caption{The figures (a) and (b) above show the test set classification accuracy for two values of N (100 and 1000 respectively) across the six preprocessing methods (start, middle, random, mean, standard deviation, and Fourier Coefficients), and three classification algorithms (SVC, RF, and KNN).  Here N represents either the number of contiguous points sampled from the start, middle, or random portions of a time series, the number of bins summary statistics were performed upon in the case of mean and standard deviation preprocessing, or the number of largest Fourier Coefficients.  In these figures, the y-axes begin at 0.4 to emphasize the relative differences between results.  A discussion of the results is given in Section \ref{ssec:results}.}
    \label{fig:bar_plots}
\end{figure*}
\section{Datasets}\label{sec:datasets}

\subsection{MIT SuperCloud Dataset}\label{ssec:mit_supercloud_dataset}
The experiments in this paper used the MIT SuperCloud Dataset~\cite{dcc}, a dataset consisting of over 2TB of data collected from the MIT SuperCloud HPC cluster.  Of primary interest to this paper, the MIT SuperCloud Dataset contains multi-channel GPU times series data consisting of features such as GPU memory utilization, GPU temperature, and GPU power draw.  Of the over 2TB of data in the MIT SuperCloud Dataset, there is approximately 65GB of labelled GPU time series data.  This labelled data consists of 3,430 manually labelled deep learning jobs which ran on the MIT SuperCloud HPC.  For further details on the MIT SuperCloud Dataset and the labelled subset see~\cite{dcc} and~\cite{wcc}. 
\subsection{Preprocessing Techniques}\label{ssec:preprocessing_techniques}
As mentioned in Section \ref{sec:introduction}, the work herein builds upon~\cite{wcc}, extending baseline implementations for workload classification.  In~\cite{wcc} the goal was to classify the different deep learning models found in the labelled dataset based upon their GPU time series.  The problem of multi-channel time series alignment was addressed in~\cite{wcc} by simply selecting the first, middle, or a random minute of data from each of the 3,430 jobs.  In this paper, we extend the results from~\cite{wcc} and introduce low overhead preprocessing techniques which address the alignment problem.
The preprocessing techniques we employ which address the alignment problem are broadly:
\begin{itemize}
\item Select a subset of N contiguous points from each time series
\item Split each full time series into N windows and perform summary statistics on each window
\item Select the N largest Fourier Coefficients for each full time series
\end{itemize}
\subsubsection{N-point Subset Selection}
This is the same technique that was used in~\cite{wcc} to address the alignment problem and is included here as a baseline for comparison with prior work.  While in~\cite{wcc} approximately 540 samples were selected from each time series, here we experimented with two sampling sizes: 100 and 1000 samples.  Furthermore, as in~\cite{wcc}, samples were taken from three different sections of each time series, the first N samples, middle N samples, and a random sampling of N points somewhere in each time series.  This resulted in six different N-point subset selection datasets (two values of N and start, middle, and random sections)
\subsubsection{N-window Summary Statistics}
One issue with N-point subset selection is that it does not consider all data in a time series.  To mitigate this issue, while addressing the alignment problem, we broke each full time series into N windows, and computed the mean and standard deviation of each window.  This resulted in a total of four N-window summary statistics datasets (two values of N and mean and standard deviation summary statistics).
\subsubsection{N-largest Fourier Coefficients}
Our third approach to the alignment problem was to select the N-largest Fourier Coefficients for each time series.  In both the discrete and continuous cases of the Fourier Transform\cite{kennedy2013hilbert,vetterli2014foundations}, the Fourier Coefficients are the components of a function (in our case GPU time series signals) projected onto a  orthogonal basis in Hilbert Space, where the basis functions or sequences are complex exponentials~\cite{kennedy2013hilbert,vetterli2014foundations}.  In this sense, selecting the N-largest Fourier Coefficients is analogous to Principal Component Analysis~\cite{jolliffe2016principal}.  In total there were two Fourier Coefficient datasets, one for each of the two values of N.

As an example, considering the SVM cluster from Figure \ref{fig:bar_plots}(a), for the first three bars, N=100 means the first 100 points, the middle 100 points, and 100 randomly sampled points respectively.  For the mean and standard deviation bars, N=100 indicates each full time series was broken into 100 windows, with statistics performed on each window.  Lastly, in the Fourier Coefficient case, N=100 indicates the dataset consisted of the largest N Fourier Coefficients.

While there are 3,430 unique jobs in the labelled dataset, as some jobs requested multiple GPUs the actual number of distinct GPU time series in our datasets was 19,481.  Each dataset was split into stratified training and testing sets using the scikit-learn StratifiedShuffleSplit class~\cite{scikit-learn}.  As the GPU portion of the labelled dataset consists of multi-channel time series dataset, we used the following sensor data collected for each GPU job:
\begin{itemize}
\item GPU Utilization (\%)
\item GPU Memory Utilization (\%)
\item Free GPU Memory (MB)
\item Used GPU Memory (MB)
\item GPU Temperature (C)
\item GPU Memory Temperature (C)
\item GPU Power Draw (Watts)
\end{itemize}
The number of time series samples for each of these seven sensors was 100 or 1000 based on the descriptions above.  Thus, taking N=100 as an example, each training dataset were in $\mathbb{R}^{15584 \times 100 \times 7}$ (15,584 trials, 100 samples per trial, seven sensors per trial).  In order to assure all data were dimensionless we applied min-max scaling, based upon the MinMaxScaler class in scikit-learn.  These scaling techniques expect a two-dimensional input so we reshaped each dataset by stacking the last two dimensions prior to scaling.  Using the example from above, the dataset after reshaping was in $\mathbb{R}^{15584 \times 700}$.  All the descriptions above also apply to the data in the test set, with the exception that the scaling models were fit only on the training data.  In total the two N values and six preprocessing techniques resulted in twelve distinct datasets which were used in our experiments.

\section{Experiments}\label{sec:experiments}

\subsection{Machine Learning}\label{ssec:machine_learning}
Using the datasets described in Section~\ref{sec:datasets}, we constructed a machine learning pipeline consisting of dimensionality reduction via PCA followed by training a classification estimator on the training data via cross-validation.  The cross-validation was done using scikit-learn's GridSearchCV~\cite{scikit-learn} with 5 cross-validation folds.  The grid search parameters for PCA were dimensions of 256 and 512.  In total, we trained three estimators (all using scikit-learn): Support Vector Classifier (SVC)~\cite{cortes1995support}, Random Forests (RF)~\cite{ho1995random,breiman2001random}, and k-Nearest Neighbors (KNN)~\cite{hastie2009elements}.  The grid search hyperparameters (naming follows the scikit-learn convention) and associated values for each of the estimators were:
\begin{itemize}
\item SVC: C (200, 300), kernel (rbf)
\item RF: n\_estimators (100, 200)
\item KNN: n\_neighbors (7, 9)
\end{itemize}
The test set was then evaluated on the best model returned by GridSearchCV and this value was used as the final accuracy in our results.
\subsection{Results}\label{ssec:results}
Experimental results are shown in Figure \ref{fig:bar_plots}.  The best test set accuracy was achieved by either the mean or standard deviation preprocessing techniques across all experiments, with the Fourier Coefficients a close second.  Of interest is the fact our proposed preprocessing techniques outperformed the start, middle, and random techniques in almost all the experiments.  While, we see the start experiments depended largely on N, the remaining five techniques were relatively independent of N.  Specifically, in all experiments the best performing model achieved an accuracy of 95\% or greater using our proposed preprocessing techniques, while the best performing baseline technique was the middle dataset, with a top accuracy of 90\%, which was stable across values of N.  It should be noted we also ran experiments for N values of 250 and 500, with results similar to those for N values of 100 and 1000.  Data and code will be made publicly available via the MIT SuperCloud Datacenter Challenge website : https://dcc.mit.edu.
\section{Conclusion}\label{sec:conclusion}

\subsection{Summary}\label{ssec:summary}
In this work we addressed the time series alignment problem via three low overhead preprocessing techniques: performing mean and standard deviation summary statistics on N bins and selecting the N largest Fourier Coefficients.  These techniques allow the alignment of time series with arbitrary length, without growing the dataset size, all while taking into account the full time series.  Additionally, given the low computational overhead of these techniques, and the high accuracy they achieve, these techniques are well suited to time series classification applications where power consumption and computational complexity are operational constraints.

\subsection{Future Work}\label{ssec:future_work}
Given the success of our proposed techniques, we envision two paths of future research on the time series alignment problem.  First, we see the investigation of kernel based methods for the classification of misaligned time series data.  This is evidenced by the fact modifications to Support Vector Machines involving dynamic time warping has been successful in this domain.  Additionally, from the point of view of a High Performance Computing operational environment, the investigation of time series preprocessing techniques to improve early classification of compute jobs is highly desirable.
\section*{Acknowledgments}
 The authors acknowledge the MIT Lincoln Laboratory Supercomputing Center (LLSC) for providing HPC resources that have contributed to the research results reported in this paper. The authors wish to acknowledge the following individuals for their contributions and support: Bob Bond, Nathan Frey, Hayden Jananthan, Tucker Hamilton, Jeff Gottschalk, Tim Kraska, Mike Kanaan, CK Prothmann, Charles Leiserson, Dave Martinez, John Radovan, Steve Rejto, Daniela Rus, Marc Zissman.
\balance
\bibliographystyle{IEEEtran}
\bibliography{IEEEabrv,references}
\end{document}